# When You Talk About "Information Processing" What Actually Do You Have in Mind?

Emanuel Diamant
VIDIA-mant, POB 933 Kiriat Ono 55100 Israel
Email: {emanl.245@gmail.com}

*Abstract*—"Information Processing" is a recently launched buzzword whose meaning is vague and obscure even for the majority of its users. The reason for this is the lack of a suitable definition for the term "information". In my attempt to amend this bizarre situation, I have realized that, following the insights of Kolmogorov's Complexity theory, information can be defined as a description of structures observable in a given data set. Two types of structures could be easily distinguished in every data set – in this regard, two types of information (information descriptions) should be designated: physical information and semantic information. Kolmogorov's theory also posits that the information descriptions should be provided as a linguistic text structure. This inevitably leads us to an assertion that information processing has to be seen as a kind of text processing. The idea is not new – inspired by the observation that human information processing is deeply rooted in natural language handling customs, Lotfi Zadeh and his followers have introduced the so-called "Computing With Words" paradigm. Despite of promotional efforts, the idea is not taking off yet. The reason – a lack of a coherent understanding of what should be called "information", and, as a result, misleading research roadmaps and objectives. I hope my humble attempt to clarify these issues would be helpful in avoiding common traps and pitfalls.

## 1. Introduction

"Information processing" is a not-so-long-ago launched buzzword that is extensively used in many research fields and communities. Despite of its widespread popularity, the real meaning of it is far less acknowledged and understood. Wikipedia [1] and Plato (Stanford Encyclopedia of Philosophy) [2] provide special entries for it, but even in the lightest manner, these entries do not confront the threatening ambiguity and incomprehensibility of this expression. Positing that "Information processing is the change (processing) of information "[1] in any way does not clarify its elusive essence. The reason for that is simple – the key component of the expression ("information") has never been defined and never determined, neither in the times of ancient philosophers nor in these glorious days, when "information era" has become our blossoming reality. It is worth to be mentioned – even today "information" does not have an accepted and a generally agreed definition. Far worse than that – it has always been (and continues to be) a "bone of contention" between many prominent thinkers, scholars and scientists.

I do not intend to take part in this controversy. In the paper's Reference section I provide a list of some relevant publications addressing this issue, with only one and a definite purpose in mind – to give the vigilant readers a fair opportunity to verify by themselves how useful and applicable are the concepts of information that these leading thinkers and scholars are developing and advance (L. Floridi [3], G. Piccinini [4], R. Capurro [5], J. Cohen [6], A. Reading [7], E. Jablonka [8], A. Sloman [9]).

To be suitable for an act of processing, information has to be something substantial. That was the reason for Michael Buckland's proposition to see information as a thing, "Information as Thing" [10]. A warm welcome despite, the idea did not survive long after its introduction.

For the mentioned above reasons, I was forced to try and to work out my own conception of "What is information?" In the rest of the paper I would like to share with you some surprising results of this my enterprise.

## 2. An intimate touch with the problem

My first encounter with information processing (problems) can be dated back to the early eighties of the past century, when, as a research engineer, I have become engaged in design and development of homeland security systems. It is well known that such systems heavily rely on visual information gathering and use. But – What is visual information? – Nobody knew then, nobody knows today. However, that has never restrained anybody from trying again and again to put up such systems and deploy them everywhere. On the other hand, I was considered that there must be a better way to cope with such a mysterious problem (as visual information processing).

I do not tend to bore you with the history of my attempts (to reach an acceptable understanding of information handling peculiarities). Interested readers are invited to visit my website {http://www.vidia-mant.info}, where a full list of my publications on the subject is available. For the sake of time and space saving, I will only provide some short excerpts from these (mostly unknown) papers.

I have dared to publish my first definition of "information" somewhere in the year 2005 [11]. At that time it has sound as follows:

Right from the beginning, it must be accepted that **information is a description**, a certain language-based description, which Kolmogorov's Complexity theory regards as a program that, being executed, trustworthy reproduces the original source object. In an image, such source objects are visible data structures from which an image is comprised of. So, a set of **reproducible descriptions of image data structures is the information contained in an image**. (Because "Visual Information" has always been my prime concern, image-related bag-of-words is ubiquitously used in my arguments. That does not mean that image-inspired definitions are only good for image information content depiction. Certainly not, certainly all definitions used for visual information content description could be easily generalized and extended to many other cases and settings).

Certainly, an image is a good example of a two-dimensional data set composed of a vast amount of closely spaced elementary picture elements (pixels). It is taken for granted that an image is not a random collection of these picture elements, but, as a rule, the pixels are naturally aggregated in specific clusters (structures). These clusters (structures) emerge as a result of data elements aggregation shaped by similarity in their physical properties (e.g., pixels' luminosity, color, brightness and so on). For that reason, I have proposed to call these structures the **primary or physical data structures**.

In the eyes of an external observer, the primary data structures are further grouped into more larger and complex aggregations, which I propose to call **secondary data structures**. These secondary structures reflect human observer's view on the arrangement of primary data structures, and therefore they **could be called meaningful or semantic data structures**. While formation of primary data structures is guided by objective (natural, physical) properties of data elements, ensuing formation of secondary structures is a subjective process guided by human habits and customs, mutual agreements and conventions.

As it has been declared earlier, **Description of structures observable in a data set has to be called "Information"**. Following the given above explanation about the nature of structures discernible in an image (in a given data set), two types of information must be distinguished therefore – **Physical Information and Semantic Information**. They are both language-based descriptions; however, physical information can be described with a variety of languages (recall that mathematics is also a language), while semantic information can be described only with the use of a natural human language.

I will not explain here what the interrelations between physical and semantic information are – Although that is a very important topic, for the purposes of this discussion, I will bring again only short excerpts from my early mentioned papers:

Every information description is a top-down evolving coarse-to-fine hierarchy of descriptions representing various levels of description complexity (various levels of description details). Physical information hierarchy is located at the lowest level of the semantic hierarchy. The process of sensor data interpretation is reified as a process of physical information extraction from the input data, followed by an attempt to associate the input physical information with physical information already retained at the lowest level of a semantic hierarchy. If such association is achieved, the input physical information becomes related (via the physical information retained in the system) with a relevant linguistic term, with a word that places the physical information in the context of a phrase which provides the semantic interpretation of it. That is, the input physical information becomes named with an appropriate linguistic label and framed into a suitable linguistic phrase (and further – in a story, a tale, a narrative), which provides the desired meaning for the input physical information. (More about the subject can be found in Diamant [12]).

## 3. From information to information processing

Now, equipped with a clear definition of "what is information", we can start to scrutinize the peculiarities of information processing procedures. Keeping in mind that physical information and semantic information are two different kinds of information, it is reasonable to investigate their processing activities separately.

Physical information processing takes place at the system's input front-end. System's input sensors (human sensing organs) constantly supply the system with huge amounts of sensor data, and this data has to be immediately processed in order to extract the physical information. (Because information processing systems are destined to process information, not data).

A common mistake is to see data processing systems as aimed to extract meaningful information from the submitted input data. Again and again – only physical information can be extracted from the data. Nothing else. (How exactly this can be done I describe in my previous papers).

As it has been already explained earlier, primary data structures are taking part in a process of further secondary structures formation. Usually, these secondary structures are the first semantic (linguistic) structures encountered in the system which constitute the lowest level of the semantic hierarchy (the first 'words' in the system). The designated words participate in the next level structure creation (e.g., a phrase or a sentence formation). The phrases are then structured in paragraphs, paragraphs in chapters, chapters in something more complex and complicated, until the whole story is accomplished.

As it was already mentioned, the rules of secondary (semantic) structures arrangement are not known in advance and not predetermined. The rules are arbitrary and subjective, established as an agreement between members of a certain user group, an outcome of their common practice and conventions. Therefore, for a successful arrangement of secondary structures the system has to be provided with a prototype semantic hierarchy, where a suitable structure is already exemplified. Traditional semantic processing architectures have also such prototyping hierarchies, but they call them with different names – e.g., previous experience records, prior knowledge databases. Their designers and users pretend that such knowledge can be derived directly from the data which is available for processing and which is representing the domain knowledge.

What I claim is that the prototypical semantic information (the prototypical semantic hierarchy) has to be provided to the system's disposal in advance, before the system starts to cope with a new task of raw data processing. And semantic information processing has to be seen as a recursive procedure of a lower level structure placement into a higher level structure, (thus the meaning, the semantics of a lower level structure is defined by its place and its use in the higher level structure).

Such search for a proper placement (of a lower level structure into a higher level one) is not always successful. In such cases, the higher level structure has to be modified to allow the accommodation of a lower level structure. That is what could be called a 'process of a new story production', a process of a prototyping information hierarchy modification and a new prototyping information hierarchy generation, which has to be seen as a semantic information processing procedure that is the basis for such cognitive tasks as reasoning, decision making, action planning, and so on.

It must be remembered that all these information processing actions are fulfilled upon linguistic text structures (compositions). It must also be remembered that text reading input systems are also processors of sensor data (visual data in text reading imaging systems, tactile data in Braille code reading systems, 0/1 sequences in electronic data handling systems). In all such cases, primary data structures are extracted first and then subjected to the lowest level semantic information processing which results with the lowest level secondary structures production (character recognition stage). The characters are then subjected to the next level semantic processing where they are composed into the first linguistic words. Only at this stage the main semantic information processing is commenced: processing of linguistic sequences, strings and pieces of text.

## 4. Computing with words

It is commonly known that semantic information processing is somehow connected with the natural language word processing custom. The paradigm "Computing with Words" (CWW) was introduced by Lotfi Zadeh in the mid-nineties of the past century [13]. Computing with Words was proposed as "a system of computation which offers an important capability that traditional systems do not have—a capability to compute with information described in natural language", [14].

It was inspired by an insight that humans perform their cognitive tasks in a very specific manner: "Computing, in its usual sense, is centered on manipulation of numbers and symbols. In contrast, computing with words, or CW for short, is a methodology in which the objects of computation are words and propositions drawn from a natural language", [15]. (To avoid any blames of misrepresentation of the core CWW principles, I will keep on to exploit extensive citations drawn from the founding fathers' seminal papers).

"Computing with words is inspired by the remarkable human capability to perform a wide variety of physical and mental tasks without any measurements and any computations. Underlying this remarkable capability is the brain's crucial ability to manipulate perceptions − perceptions of distance, size, weight, color, speed, time, direction, force, number, truth, likelihood and other characteristics of physical and mental objects. Manipulation of perceptions plays a key role in human recognition, decision and execution processes. As a methodology, computing with words provides a foundation for a computational theory of perceptions… A basic difference between perceptions and measurements is that, in general, measurements are crisp whereas perceptions are fuzzy..." [15].

Thus, Computing with words assumes that "computers would be activated by words, which would be converted into a mathematical representation using fuzzy sets (FSs), and that these FSs would be mapped by means of a CWW engine into some other FS, after which the latter would be converted back into a word "[16].

"Another basic assumption in CW is that information is conveyed by constraining the value of variables. Moreover, the information is assumed to consist of a collection of propositions expressed in a natural or synthetic language, that is, variables take as possible values linguistic ones" [17].

"One objective of Computing with Words is to enable the inclusion of human sourced information in the formal computer based decision-making models that are becoming more and more pervasive. Central to CWW is a translation process. This process involves taking linguistically expressed information and translating into a machine manipulative format. The types of information that have to be translated are not restricted to the linguistic values of
variables but must also include linguistically expressed information for processing information", [18].

"Another objective of CWW is to help in the human understanding of the results of information acquisition and information processing. This involves techniques of linguistic summarization and retranslation. Retranslation involves taking the results of the manipulation of formal objects and converting them into linguistic terms understandable to the human. Here we are going in the opposite way of the previous objective. With linguistic summarization we are trying to summarize large sets of data, with the aid of words, in a way that a human can get a global understanding of the content of the data", [18].

"The use of the linguistic semantic model based on type-2 fuzzy sets is a current trend in decision making. Several recent works have developed new decision making models in which the linguistic information is computed and aggregated by means of interval type-2 fuzzy sets to maintain a higher (and more realistic) degree of uncertainty of the linguistic information", [17].

Initially, CWW has been accepted with great admiration and great expectations have been aroused when attempts to apply CWW principles to different aspects of human information processing customs have been considered, e.g., perception computing and judgment [19], reasoning [20], decision making [17], [21], and text processing [22]. Unfortunately, these expectations have not been satisfied. "The theory of CW, as currently introduced, is considered raw and needs intensive work and research before it can be applied to the practical use. Since its introduction, quite a few researches have been undertaken in this area but the ultimate goal of building an actual CW computational engine has thus far proven to be elusive", [20].

"It is important not to confuse CW with natural language processing. CW does not claim that it is able to fully model complex natural language propositions nor does it argue that it can perform reasoning on such statements. But it rather offers a system of computation that is superior to the traditional bivalent computing systems because of its capability to reason and compute with linguistic words hence modeling human reasoning ", [20].

To summarize, in this brief review of the CWW literature one thing must be noted and not to be left unattended: dealing with information representation issues, CWW never asked itself the question: What is information? And then, as a consequence, dealing with undefined (linguistic) information, CWW repeatedly associates it with a single word or with a bundle of several "precisiated" words. This, naturally, leads it to difficulties and troubles, some of which have been just mentioned above.

## 5. Concluding remarks

What follows from the proposed definition of information (as a linguistic description of structures observable in a data set) is that information processing must be defined as a text processing enterprise. And not as an act of processing of separate single words or simple word compositions, like it is commonly done in the CWW paradigm, ontology-based world representations, key-words data mining (practice), and so on.

How this speculative declaration can be converted into a practical implementation? – I still do not know (at least at the current stage of my research). What I do know and about what I am perfectly certain is that the term "computation" is not applicable to the action that hypothetically takes place when humans are busy with processing information, that is, are busy with processing linguistic texts.

I hope that my humble insights about the essence and the linguistic nature of the term "information" would be helpful in paving the way to an increased information-related issues appreciation and establishing the right means for an effective information processing accomplishment.